\documentclass[10pt,twocolumn,letterpaper]{article}

\usepackage{cvpr}
\usepackage{times}
\usepackage{epsfig}
\usepackage{graphicx}
\usepackage{subfig}
\usepackage{amsmath}
\usepackage{amssymb}
\usepackage{booktabs}

\usepackage{algorithm}
\usepackage{algorithmic}

\usepackage[pagebackref=true,breaklinks=true,letterpaper=true,colorlinks,bookmarks=false]{hyperref}

\cvprfinalcopy % *** Uncomment this line for the final submission

\def\cvprPaperID{2736} % *** Enter the CVPR Paper ID here

\ifcvprfinal\fi
\begin{document}

%%%%%%%%% TITLE
\title{Distribution Networks for Open Set Learning}

\author{Chengsheng Mao, Liang Yao,  Yuan Luo\\
Northwestern University\\
Chicago, IL, 60611\\
{\tt\small \{chengsheng.mao, liang.yao, yuan.luo\}@northwestern.edu}
% \and
}

\maketitle
%\thispagestyle{empty}

%%%%%%%%% ABSTRACT
\begin{abstract}
   In open set learning, a model must be able to generalize to novel classes when it encounters a sample that does not belong to any of the classes it has seen before. Open set learning poses a realistic learning scenario that is receiving growing attention. Existing studies on open set learning mainly focused on detecting novel classes, but few studies tried to model them for differentiating novel classes. In this paper, we recognize that novel classes should be different from each other, and propose distribution networks for open set learning that can model different novel classes based on probability distributions. We hypothesize that, through a certain mapping, samples from different classes with the same classification criterion should follow different probability distributions from the same distribution family. A deep neural network is learned to map the samples in the original feature space to a latent space where the distributions of known classes can be jointly learned with the network. We additionally propose a distribution parameter transfer and updating strategy for novel class modeling when a novel class is detected in the latent space. By novel class modeling, the detected novel classes can serve as known classes to the subsequent classification. Our experimental results on image datasets MNIST and CIFAR10 show that the distribution networks can detect novel classes accurately, and model them well for the subsequent classification tasks.
\end{abstract}

%%%%%%%%% BODY TEXT
\section{Introduction}

In traditional image classification tasks, all classes in the test set are restricted to the classes seen before in the training set. This is referred to as close set learning (CSL). Although a vast majority of recognition systems are designed for close set and substantial progress have been made in CSL, it requires that all classes are known in advance and there are enough samples from each class in the training set, which may not be a realistic setting because one can not always know all possible classes beforehand. A novel object not belonging to any of the classes in the training set will cause a failure to a CSL system.

Human can easily learn new concepts when seeing a novel object, e.g., when a child who only knows dogs and cats sees a whale for the first time, he knows it is different from what he has seen before, and there should be a concept in his mind about this animal, thus, whenever he meets a whale afterwards, he can easily know this is of the new kind that he has seen. Though learning new concepts is easy for humans, it is a challenge for an artificial intelligence (AI) system \cite{dietterich2017steps}. This motivates us to study open set learning (OSL) that can model novel classes not in the training set.

In general, we require a model that can handle OSL problems to have the following properties: (1) it should be trainable from a limited number of samples from a limited number of known classes; (2) it can detect samples from unknown classes; (3) if a sample is detected from an unknown class, it can model the class and make this class known to the subsequent classification; (4) if a sample is detected from a known class, it can accurately classify this sample to its true class; (5) as the number of known classes grows, its computational requirement and memory footprint should remain bounded, or at least grow very slowly. Although computer vision has made great progress in image classification tasks, to the best of our knowledge, there is not a single algorithm satisfying all of the above requirements for open set image classification. Most existing studies on OSL can detect but can not model novel classes for subsequent classification, thus violate (3).

\begin{figure*}
  \centering
  \includegraphics[width=\textwidth,height=0.3\linewidth]{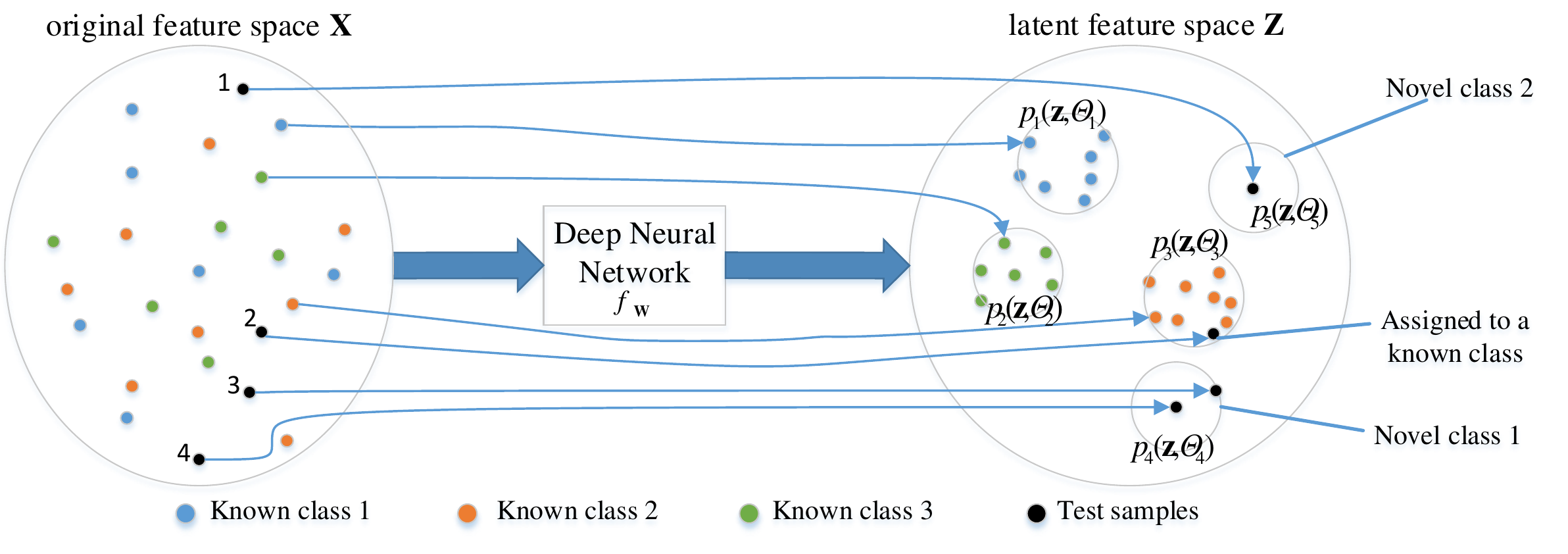}
  \caption{The architecture of Distribution Networks. Samples in the original feature space \textbf{X} are mapped to a low-dimensional latent feature space \textbf{Z} by a deep neural network $f_\mathbf{W}$. In the latent space \textbf{Z}, each class follows a distinctive probability distribution $p(\mathbf{z},\mathbf{\Theta})$ with learnable parameter(s) $\mathbf{\Theta}$. If a test sample is detected in an unknown class, then a novel class is generated and modeled with a new probability distribution $p_{new}(\mathbf{z},\mathbf{\Theta}_{new})$ for the test sample, e.g., new distribution $p_{4}(\mathbf{z},\mathbf{\Theta}_{4})$ is generated for test samples 3 and 4 in novel class 1, and $p_5(\mathbf{z},\mathbf{\Theta}_{5})$ is generated for test sample 1 in novel class 2. If a test sample is not novel, e.g., test sample 2, it is classified to a known class with the highest posterior probability as a standard close set classifier does.}
    \label{architecture}
\end{figure*}

In this paper, we propose a model that can handle OSL problems well. In our views, different classes that can be compared to each other usually have some features in common and have distributions of a common type. There should be some discriminative differences in feature values and distribution parameters with respect to the different classes. From this view, we assume objects from different classes should have certain discriminative features that follow different distributions from the same distribution family. For OSL, we assume the distributions of all classes (known and unknown) have the same distribution types but different parameter values. Thus, if a test sample is detected from an unknown class, we generate a distribution corresponding to this novel class by parameter transfer. The generated class can be regarded as known for the classification of next samples. So repeatedly, by the end of the test process, each test sample is classified to an original known class or to a newly created class.

We named our model Distribution Networks (DN). The key insight of DN is to map the original high-dimensional space where the class distributions are unknown and difficult to estimated to a low-dimensional latent space where the distribution can be easily estimated. The model is illustrated in Figure \ref{architecture}. It consists of two parts, the first part is a deep neural network for feature learning, the target is to learn representative features that can be used to construct discriminative distributions between different classes; the second part is an open set classifier based on probability distribution in the latent feature space.

The novelty of our work is as follows: (1) At the modeling level, we propose Distribution Networks to map complex high-dimensional distributions to low-dimensional ones that are expected to be discriminative to different classes including unknown classes. (2) At the problem level, we redefine the open set learning problem, since few studies on this problem can model the novel classes to make them known to subsequent test samples, we focus on the novel class modeling sub-problem, and try to tackle it by Distribution Networks in this work.

\section{Related Work}
There are several related lines of research that have connections to our work. We briefly outline their connections and differences with our work.

\textbf{Open Set Recognition.}
Since Scheirer et al. \cite{scheirer2013toward,scheirer2014probability} defined the "open set recognition problem", more and more researchers have made their effort to this realistic problem \cite{bendale2015towards,bendale2016towards,gunther2017toward,jain2014multi,cardoso2015bounded}. These efforts have made significant contributions to the development of OSL, however, they did not model the novel classes for subsequent classification which is important for a robust AI system to learn novel knowledge. In our work, we recognize the differences between novel classes, and propose and implement an algorithm to model novel classes for subsequent classification.

\textbf{Zero-shot Learning.}
Zero-shot learning \cite{socher2013zero,palatucci2009zero,zhang2017learning,kodirov2017semantic} and OSL share the feature that there are no training data for some classes in test data. However, zero-shot learning is often regarded as CSL because all classes are known before test, and there usually are semantic descriptions for the no-show classes in training set for classification. While in OSL, we discover new classes and model them in the test process. Though the tasks are different, some idea for zero-shot learning can also be leveraged for OSL. The idea in the work \cite{socher2013zero} for zero-shot learning is somewhat similar to ours. The differences are that they project images into a semantic word space where all classes including the no-show classes in training set have word vectors for training, while in OSL, we know nothing about the unknown classes before seeing a sample that is abnormal to all known classes, the only information is that unknown classes should have different distributions from known classes.

\textbf{Non-exhaustive Learning.}
Non-exhaustive learning tries to learn from a non-exhaustive training set where some classes are missing. Murat et al. \cite{Dundar2009Learning} proposed an empirical Bayesian approach to deal with non-exhaustive training datasets. Ferit et al. \cite{akova2010machine} assumed Gaussian distributions for all classes (known and unknown) and dynamically updated class set based on the maximum likelihood detection of novelties. Murat et al. \cite{dundar2012bayesian} defined a Dirichlet process prior model over class distributions to ensure that both known and unknown class distributions originate according to a common base distribution. However, these methods may have difficulties in high dimensional image data. DN proposed in this paper can deal with high dimensional or unstructured data naturally by learning a low-dimensional representation.

\textbf{Few-shot Learning.}
Few-shot learning including one-shot learning, tries to do learning tasks by using very few training examples \cite{koch2015siamese,vinyals2016matching,snell2017prototypical}. In OSL, when a novel class is generated, there should be very few samples in the novel class, the subsequent classification for samples in this class can be seen as few-shot learning. In our work, the model of novel class is learned from the known classes and the limited samples. The prototypical networks \cite{snell2017prototypical} for few-shot learning has something in common with ours. We both map the original feature space to an embedding space. The prototypical networks try to figure a prototype for each class in the embedding space for few-shot learning, while our DN try to model a distribution for each known class and to transfer the distribution parameters to unknown classes for open set learning.

\textbf{Class Incremental Learning.}
Class incremental learning tries to train a model incrementally with new training data including new classes \cite{kemker2018fearnet,gepperth2016bio,rebuffi2017icarl}, that is, samples of new classes with labels are added to train an updated model after a learning system being trained. Open set learning have some similar features with class incremental learning. The two learning tasks both try to update a learned model for classifying samples from novel classes. However, instead of updating a learned model in retraining sessions with known samples from novel classes, our open set learning framework updates the learned model in the test process when a sample from unknown classes is detected. Thus, In addition to modeling the known classes in the training set, open set learning can also model the novel classes appeared in the test process.

\textbf{Novel Class Detection.}
Novel class detection is an essential part of OSL. Most studies on novel class detection \cite{Masud2011Classification,da2014learning,al2016recurring,mu2017streaming,haque2016sand} try to train a classification model that can detect unknown classes while still generalizing to new instances of known classes. Studies applying deep neural network for novel image class detection include \cite{neal2018open,Ge2017GenerativeOF,hassen2018learning}. However, these studies can detected novel classes but can not model them for subsequent classification. The distribution network proposed in this paper can detect novel classes and model them.

\section{Methods}
\subsection{Problem Description}
In the initial training set, there are a number of labeled samples from $K$ known classes $\{1,\ldots,K\}$. Let $S_k$ denotes the set of samples labeled with class $k$, i.e., $S_k=\{\mathbf{x}_1^k,\ldots,\mathbf{x}_{n_k}^k\}$, where $\mathbf{x}_i^k\in \mathbb{R}^D$ is the $D$-dimensional feature vector from class $k$ in the original feature space, $n_k$ is the number of samples in class $k$. Given a sequence of test samples $Te=\{\mathbf{x}_1,\ldots,\mathbf{x}_n\}$, the task is to label the samples with  known class $\{1,\ldots,K\}$ or learned novel classes $\{K+1,\ldots,K+m\}$.

\subsection{Distribution Networks} \label{sec:DN}
Distribution Networks map samples in the $D$-dimensional original space $\mathbb{R}^D$ to their representations in an $M$-dimensional latent space $\mathbb{R}^M$ such that they follow a certain type of probability distribution in the latent space. We define the mapping as a deep neural network $f_\mathbf{W}:\mathbb{R}^D\rightarrow\mathbb{R}^M$ with learnable parameters $\mathbf{W}$. In the latent space $\mathbb{R}^M$, we assume samples from class $k$ follow a probability distribution $p_k(\mathbf{z};\mathbf{\Theta}_k)$ with learnable parameters $\mathbf{\Theta}_k$. In the following settings, we denote the embedding of $\mathbf{x}_i^k$ in the latent space as $\mathbf{z}_i^k$, i.e., $\mathbf{z}_i^k=f_\mathbf{W}(\mathbf{x}_i^k)$.

Our training objective is to make samples more likely to belong to their labeled class. Thus, we try to maximize the log likelihood of each class with respect to their samples. For class $k$, the log likelihood is
\begin{equation}\label{eq:loglikelihood}
\log\mathcal{L}_k(\mathbf{\Theta}_k, \mathbf{W}) = \sum_{i=1}^{n_k}\log p_k(\mathbf{z}_i^k;\mathbf{\Theta}_k)
\end{equation}

To balance the effect of each sample on the results and prevent classes with more samples from outweighing classes with less samples, we use the negative mean log likelihood for each class in the final loss function, thus for a training set with known classes $\{1,\cdots,l\}$, the loss function is defined as Eq. \ref{eq:loss}, where $p_k(\cdot;\mathbf{\Theta}_k)$ is a specified probability density function with parameters $\mathbf{\Theta}_k$ for class $k$. Learning proceeds by minimizing $J(\mathbf{W},\mathbf{\Theta})$ via stochastic gradient descent (SGD) \cite{sutskever2013importance} for each epoch.
\begin{equation}\label{eq:loss}
\begin{aligned}
J(\mathbf{W},\mathbf{\Theta}) & = -\sum_{k=1}^l \frac{1}{n_k}\log\mathcal{L}_k(\mathbf{\Theta}_k, \mathbf{W})   \\
&=-\sum_{k=1}^l \frac{1}{n_k}\sum_{i=1}^{n_k}\log p_k(f_\mathbf{W}(\mathbf{x}_i^k); \mathbf{\Theta}_k)
\end{aligned}
\end{equation}

In this paper, we assume that each class follows a multivariate Gaussian distribution in the latent space, thus $p_k(\mathbf{z};\mathbf{\Theta})$ represents the Gaussian probability density function (Eq. \ref{eq:Gaussian}) and $\mathbf{\Theta}_k$ consists of the center $\mu_k$ and the covariance matrix $\Sigma_k$.

\begin{equation}\label{eq:Gaussian}
p_k(\mathbf{z};\mu_k,\Sigma_k)=\frac{\exp(-\frac{1}{2}(\mathbf{x}-\mu_k)^T\Sigma_k^{-1}(\mathbf{x}-\mu_k))}{\sqrt{(2\pi)^k|\Sigma_k|}}
\end{equation}

Due to the unimodal feature of Gaussian distribution, samples from a class tend to be mapped to a cluster around its center, $\Sigma$ defines the variance of the cluster. Through minimizing the loss defined in Eq. \ref{eq:loss}, a class will become more compact, and thus different classes are more discriminative. Though the mapping network is learned based on known classes, we expect the learned mapping can also compact novel classes.

To achieve this expectation, we assume that there exists a feature set $Z=f(X)$ such that the distributions of all classes on $Z$ have the same distribution type (assumed Gaussian in this paper) with different parameters. This implies that the feature set $Z$ can discriminate all classes (known and unknown). We aim to learn this feature set from the original features $X$ using Distribution Networks based on the known classes.

\subsection{Validation Strategies}
As the training progresses, the classes become increasingly compact. To prevent training an over-fitting model, in the training process, we evaluate the discriminability between classes in an independent validation set. The model with the highest discriminability in the validation set is selected. To simulate the open set environment, the validation set contains a couple of unknown classes that are not presented in the training set as well as the known classes. Let the training set contains $l$ known classes $\{1,\cdots,l\}$, and the validation set contains $K$ classes $\{1,\cdots,K\}$ where classes $\{l+1,\cdots,K\}$ are absent in training process. The evaluation of a model is based on samples in the validation set. Classification performance of known classes is evaluated based on classes $\{1,\cdots,l\}$ and performance of unknown class is evaluated based on classes $\{l+1,\cdots,K\}$.

\textbf{Posterior distribution estimation.}
We consider the learned probability distribution function $p_k(\mathbf{z}; {\mu}_k, \Sigma_k)$ for class $k\in\{1,\cdots,l\}$ as the prior distribution, thus the conjugate prior of $(\mu, \Sigma)$ can be modeled as a Normal-inverse-wishart (NIW) distribution. Thus, the posterior of $(\mu, \Sigma)$ is also a NIW distribution with parameters updated by Eq. \ref{eq:posterior}, where $\overline{\mathcal{D}}$ and $N$ are the mean and the sample number of data $\mathcal{D}$, respectively, and $S_\mathcal{D}=\sum_{\mathbf{x}\in\mathcal{D}}{\mathbf{x} \mathbf{x}^T}$. Thus, $\mu$ and $\Sigma$ can be estimated as Eq. \ref{eq:MAPestimation}  by maximum a posteriori (MAP) estimation \cite{murphy2012machine}, where $D$ is the dimension of the data.
\begin{equation}\label{eq:posterior}
\begin{aligned}
p(\mu,\Sigma |\mathcal{D}) &=NIW(\mu,\Sigma | m_N,\kappa_N,\nu_N,S_N)  \\
\kappa_N &=\kappa_0+N  \\
\nu_N &=\nu_0+N  \\
m_N &= (\kappa_0m_0+N\overline{\mathcal{D}})/\kappa_N  \\
S_N &= S_0+ S_\mathcal{D}+\kappa_0 m_0 m_0^T-\kappa_N m_N m_N^T
\end{aligned}
\end{equation}

\begin{equation}\label{eq:MAPestimation}
\hat{\mu} = m_N; \hat{\Sigma}=S_N/(\nu_N +D+2)
\end{equation}

In our settings, we try to use the validation set to update the learned distribution. For each known class, $\kappa_0$ and $\nu_0$ are the number of samples in this class in the training set, $m_0$ and $S_0$ are the center and covariance matrix joint learned by the network, respectively. For unknown classes in the validation set, since there are no data of this class available in the training set, all the original parameters (i.e., $\kappa_0$, $\nu_0$, $m_0$ and $S_0$) are set to 0. Data $\mathcal{D}$ is the sample in the corresponding class in the validation set. Thus, we get a Gaussian distribution $p_k(\mathbf{z};\hat{\mu}_k, \hat{\Sigma}_k)$ for each class $k\in\{1,\cdots,l\}$.

\textbf{Model selection.}
Given a learned mapping network $f_\mathbf{W}(\cdot)$, the likelihood of a sample $\mathbf{x}$ belonging to class $k$ is $p_k(f_\mathbf{W}(\mathbf{x});\hat{\mathbf{\Theta}}_k)$ where $\hat{\mathbf{\Theta}}_k$ consists of $\hat{\mu}_k$ and $\hat{\Sigma}_k$. We define a threshold $t_k$ for each class $k$ and compute its F1 score as its discriminability. A sample $\mathbf{x}$ is considered not belonging to class $k$ if $p_k(f_\mathbf{W}(\mathbf{x}),\mathbf{\Theta}_k) < t_k$. For class $k$, the F1 score is computed based on all samples in the validation set and the threshold $t_k$. As $t_k$ balances the trade-off between recall and precision for a higher F1 score, we select the threshold $t_k$ that achieves the highest F1 score for class $k$ on the precision-recall curve. In different training epochs, We use the weighted sum of the F1 scores of all classes $D = \sum_{i=1}^K \lambda_i F_i$ as the metric to select the model,
% \begin{equation}\label{eq:discriminability}
% D = \sum_{i=1}^K \lambda_i F_i
% \end{equation}
where $F_i$ and $\lambda_i$ denote the F1 score and weight of class $i$, respectively. $\lambda_i$ is user-assigned to make some classes more or less discriminable. Generally, increasing the weight of novel classes ($\{l+1,\cdots,K\}$) will favor a model discriminative to unknown classes. The algorithm of training and validation is described in Algorithm \ref{al:train} where PRCurve() in Line \ref{line:prcurve} computes the precision-recall curve with different thresholds, and returns Precisions (P), Recalls (R), F1-scores (F) and Thresholds (T).

Though other metrics to determine the divergence between two distributions can be used to define the discriminability between two classes, the discriminability among multiple distributions poses difficulty for this alternative formulation. Because the test process requires a threshold $t_k$ for class $k$ to decide whether to reject a sample or not, to make validation and test conditions match, we design that the validation process also discriminate samples based on a threshold $t_k$ for class $k$. Additionally, By defining the discriminability as the highest F1 scores, the corresponding thresholds can be directly transferred to the test process.

In summary, the validation process provides two advantages for the learning process: (1) it can prevent overfitting by evaluation a learned model on the validation set. (2) it can estimate the distribution parameters ($\mathbf{\Theta}$ and $t$) for all classes which are essential for novel class detection and modeling in the test process.

\begin{algorithm}[t]  \footnotesize
\caption{  Distribution Networks Training and validation. }
\label{al:train}
\begin{algorithmic}[1]
\REQUIRE ~~ \\
$Tr_k=\{\mathbf{x}_{tr_1}^k,\ldots,\mathbf{x}_{tr_{n_k}}^k\}, 1\leq k\leq l$, training set; \\
$V_k=\{\mathbf{x}_{v_1}^k,\ldots,\mathbf{x}_{v_{m_k}}^k\}, 1\leq k\leq K$, validation set; \\
$\lambda_k, 1\leq k\leq K$, the discrimination weight; \\
$epochs$, the total training epochs.  \\
\ENSURE   ~~ \\
$f_\mathbf{W^s}$, the selected model;  \\
$t_k^s, 1\leq k\leq K$, the threshold;  \\
$\mathbf{\Theta}_k^s, 1\leq k\leq K$, the distribution parameters $\mu_k, \Sigma_k$.

\STATE Initialize network parameters $\mathbf{W}$ and $\mathbf{\Theta}_k, 1\leq k \leq l$.
\STATE  $i=0$,  $D_{best}=0$
\WHILE{ epoch $i < epochs $}
\STATE Learn a network $f_\mathbf{W}$ and $\mathbf{\Theta}_k$ based on $Tr_k$,  $1\leq k \leq l$.
\STATE $\mathbf{z}_{v_j}^k = f_\mathbf{W}(\mathbf{x}_{v_j}^k), 1\leq j \leq m_k, 1 \leq k \leq K $
\FOR { $k$ in $\{1,\cdots,K\}$ }
\STATE Update $\mathbf{\Theta}_k$ with Eq. \ref{eq:posterior} and \ref{eq:MAPestimation}
\STATE $l_{v_j}^{s\rightarrow k} = p_k(\mathbf{z}_{v_j}^s,\mathbf{\Theta}_k) ,  1\leq j \leq m_s, 1 \leq s \leq K$
% \FOR {threshold $t \in \{l_{v_j}^{s\rightarrow k} | 1\leq j \leq m_s, 1 \leq s \leq K\}$}
% \STATE Compute the precision, recall and f1 score.  \\
% $precision = \frac{N(\{ l_{v_j}^{k\rightarrow k} \mid l_{v_j}^{k\rightarrow k}  \geq t, 1\leq j \leq m_k \})}{N(\{ l_{v_j}^{s\rightarrow k} \mid l_{v_j}^{s\rightarrow k} \geq t, 1\leq j \leq m_s, 1 \leq s \leq K \})}$;
% $recall = \frac{N(\{ l_{v_j}^{k\rightarrow k} \mid l_{v_j}^{k\rightarrow k}  \geq t, 1\leq j \leq m_k \})}{m_k}$
% $f1(t) = \frac{2*precision*recall}{precision+recall}$
% \ENDFOR
\STATE P, R, F, T = PRCurve($\{l_{v_j}^{s\rightarrow k} | 1\leq j \leq m_s, 1 \leq s \leq K\}$)  \label{line:prcurve}\\
$idx=\arg \max F$, $F_k = F[idx], t_k = T[idx]$
\ENDFOR
\STATE $D = \sum_{i=1}^K \lambda_i F_i$
\IF {$D > D_{best}$}
\STATE $f_{\mathbf{W}^s} = f_\mathbf{W}$ ; $t_k^s = t_k$;  $\mathbf{\Theta}_k^s = \mathbf{\Theta}_k$
\ENDIF
\STATE $i=i+1$
\ENDWHILE
\RETURN
\end{algorithmic}
\end{algorithm}

\subsection{Test on Open Set}
As we get the mapping network $f_\mathbf{W}$ and the probability distribution function $p_k(\cdot)$ together with the threshold $t_k$ of each known class $k$ from the validation process, given a test sample $\mathbf{x}$, we first detect if it is from a novel class, if so, we should model a new class for it; if not, the standard close set classification process (with all known and novel classes accumulated so far) is executed.

\begin{algorithm}[t] \footnotesize
\caption{Testing algorithm on open set. }
\label{al:test}
\begin{algorithmic}[1]
\REQUIRE ~~  \\
$Te=\{\mathbf{x}_1,\ldots,\mathbf{x}_n\}$, test set;  \\
$f_\mathbf{W_s}$, the selected model; \\
$t_k^s, 1\leq k\leq K$, the threshold for each class $k$;\\
$\mathbf{\Theta}_k^s, 1\leq k\leq K$, the distribution parameters $\mu_k, \Sigma_k$.

\ENSURE  ~~\\
The label $y_i$ for each sample $\mathbf{x}_i$ in the test set;  \\
the class set $C$ in test set;    \\
the distribution parameters $\mu_k$ and $\Sigma_k$ for each class $k$; \\
the thresholds $t_k$ of for each novel class $k$.

\STATE Initialization. $C=\{1,\cdots,K\}$, $I=K+1$
\FOR  {$\mathbf{x}_i$ in $Te$}
  \STATE  $\mathbf{z}_i = f_{\mathbf{W}^s}(\mathbf{x}_i)$, $l_k = p_k(\mathbf{z}_i,\mathbf{\Theta}_k^s), k\in C $
  \IF {$l_k<t_k^s$ for all $k \in C$}
    \STATE $y_i=I$  \qquad  // \ assign the sample a new class label
    \STATE Parameter transfer using Eq. \ref{eq:gaussianparameters}
    \STATE $\kappa_I=1, \nu_I=1$
    \STATE $C= C\bigcup \{I\}, \qquad I=I+1$
  \ELSE
  \STATE $y_i = \max_{k\in C} l_k, \quad s.t. \quad  l_k \geq t_k^s$
  \STATE $\Sigma_{y_i}= \frac{\Sigma_{y_i}(\nu_{y_i}+D+2) +  \frac{\kappa_{y_i}}{\kappa_{y_i}+1}(\mathbf{z}_i-\mu_{y_i})(\mathbf{z}_i-\mu_{y_i})^T}{\nu_{y_i}+D+3} $;  \\
  $\mu_{y_i}=(\kappa_{y_i}\mu_{y_i}+\mathbf{z}_i)/(\kappa_{y_i}+1)$; \\
  $\kappa_{y_i}=\kappa_{y_i}+1$, $\nu_{y_i}=\nu_{y_i}+1$
  \ENDIF
\ENDFOR
\RETURN
\end{algorithmic}
\end{algorithm}

\textbf{Novel class detection.}
For a test sample $\mathbf{x}$, the likelihood of it belonging to class $k$ is computed as $p_k(f_\mathbf{W}(\mathbf{x}),\mathbf{\Theta}_k)$. We consider the likelihood for predicting whether this sample is from a known class. The threshold $t_k^s$ selected in the validation process comes in handy for class $k$  to reject a sample if $p_k(f_\mathbf{W}(\mathbf{x}),\mathbf{\Theta}_k) < t_k^s$. If a sample is rejected by all the known classes, it is considered to be from a novel class. Otherwise, it is considered to be from a known class that accepts it with the highest likelihood. By making this decision, we take the underlying assumption that all known classes have an equal prior probability.

\textbf{Novel class modeling.}
Most of the existing open set recognition algorithms only detect if a sample is novel, while never differentiate or model the novel classes. We believe that modeling the novel classes can help discover new patterns and even new knowledge in the real world. We try to discover and model the novel classes in the test process and make it known to the subsequent classification.

In the test process, if a sample is rejected by all the known classes, we model a novel distribution for the class of this sample based on the information of known classes. Because the novel class are absent in the training set just like some validation classes (i.e., class $\{l+1,\cdots,K\}$), we transfer the parameters of these classes to the novel class. Therefore, for a sample $\mathbf{z}$ from a novel class $I$, the novel class is modeled as a Gaussian distribution with center $\mu_{I}$ and covariance matrix $\Sigma_{I}$, and the threshold $t_{I}$ is also transferred. The parameter transfer strategy is defined as Eq. \ref{eq:gaussianparameters}, where $\Sigma_i$ and $t_i$ are estimated in the validation process. After a distribution of a novel class is created, the class is marked known for the subsequent classification.
\begin{equation}\label{eq:gaussianparameters}
\begin{aligned}
\mu_{I} &= \mathbf{z};   \\
\Sigma_{I} &= \frac{1}{K-l}\sum_{i=l+1}^K \Sigma_i; \\
t_{I} & = \frac{1}{K-l}\sum_{i=l+1}^K t_i
\end{aligned}
\end{equation}

In the subsequent classification process, if a sample is assigned to a known class $k$, we update the distribution parameters $\mu_{k}$ and $\Sigma_{k}$ based on the newly assigned sample. We consider the current Gaussian distribution as the prior distribution of the class and using the NIW distribution as the conjugate prior of $\mu_{k}$ and $\Sigma_{k}$. We can also update the distribution parameters $\mu_{k}$ and $\Sigma_{k}$ base on Eq. \ref{eq:posterior} and \ref{eq:MAPestimation} with $N=1$ for a test sample. Our algorithm for test on open set can be described in Algorithm \ref{al:test}.

One may want to model new classes by clustering the detected novel samples. However, in practice, the test samples are not always obtained at the same time. Incremental test settings in our test method are more realistic, where samples are fed to a model for test one by one, i.e., the samples coming first do not know anything about the later samples. Thus, if a sample is detected as in an unknown class, modeling a novel class for this sample should be based on the known information and not on the subsequent samples.

\subsection{Scalability Analysis}
After the mapping network is trained. To classify a sample, we only need to map the sample to the latent space and then compute the likelihood of this sample in each known classes. The computational complexity is $O(C)$ for classifying a new sample, where $C$ is the number of current classes. As for the memory footprint, besides the network parameters, the system also need to store the parameters ($\mu$, $\Sigma$, $t$ and $\kappa$) for each known class, thus the space complexity is $O(C)$. To add a new class to the system, we only need the memory to store the four parameters instead of storing a large number of samples. Thus, our model can be scalable with the number of classes.

\begin{table*}[t] \footnotesize
\centering
  \caption{Open set classification F1 score on MNIST dataset. The best performances are bolded. DN=Distribution Networks. $D$ is the output dimension. `mean' is the average F1 score for all the unknown classes. `one unknown' consider all unknown classes as a class and compute the F1 score.}
    \begin{tabular}{l|c|cccc|c}
    \toprule
    Models   & \multicolumn{1}{c|}{known classes} &   \multicolumn{4}{c}{unknown classes} \\
    \cline{3-7}
          &  F1-micro   &     7 (validation)   &   8    & 9   & mean   & \multicolumn{1}{c}{one unknown}  \\
          \midrule
    OpenMax &   0.958 &     -      &   -      &    -   & -     & 0.505  \\
    SoftMax &   0.951 &    -      &   -    &   -  & -  & 0.476  \\
    One-vs-Set &  0.754 &     -      &   -      &    -  & -     & -  \\
    WSVM &   0.699 &     -      &   -      &    -   & -     & -  \\
    PI-OSVM &   0.565 &     -      &   -      &    -   & -     & -  \\
    PI-SVM  &   0.855 &     -      &   -      &    -     & -   & -  \\
    DN,Isometric,D=100 &  0.908$\pm${0.050} & 0.667$\pm${0.072} & 0.458$\pm${0.100} & \textbf{0.803}$\pm${0.013} & 0.643$\pm$0.042 & 0.840$\pm${0.059} \\
    DN,Share,D=10 & 0.832$\pm$0.050 &	0.638$\pm$0.036	&	\textbf{0.691}$\pm$0.039	&	0.786$\pm$0.038 &	0.705$\pm$0.024	&	0.744$\pm$0.050 \\
    DN,Share,D=100 &  0.966$\pm$0.001 & 0.639$\pm$0.063 & 0.491$\pm$0.056	&	0.664$\pm$0.020 &	0.598$\pm$0.031	&	0.917$\pm$0.002 \\
     DN,Share,D=50 & \textbf{0.973}$\pm$0.001 &	\textbf{0.768}$\pm$0.021	&	0.597$\pm$0.048	&	0.756$\pm$0.009	& \textbf{0.707}$\pm$0.020	&	\textbf{0.937}$\pm$0.004 \\
    DN,ShareDiag,D=10 &  0.968$\pm$0.002 &	0.479$\pm$0.067	&	0.621$\pm$0.044	&	0.690$\pm$0.058	& 0.597$\pm$0.017	&	0.930$\pm$0.005  \\
    DN,ShareDiag,D=100 &  0.930$\pm$0.029 &	0.542$\pm$0.089	&	0.463$\pm$0.075	&	0.722$\pm$0.021	& 0.575$\pm$0.041	&	0.872$\pm$0.043 \\
    \bottomrule
    \end{tabular}  \\%https://www.overleaf.com/project/5be9e4599f0dd86be9c51365
    OpenMax is in \cite{bendale2016towards}; One-vs-Set is in  \cite{scheirer2013toward}; WSVM is in \cite{scheirer2014probability}; PI-OSVM and PI-SVM are in \cite{jain2014multi}
  \label{tab:mnist}%
\end{table*}%

\begin{table*}[t] \footnotesize
  \centering
  \caption{Open set classification F1 score on CIFAR10 dataset. The best performances are bolded. DN=Distribution Networks. $D$ is the output dimension. `mean' is the average F1 score for all the unknown classes. `one unknown' considers all unknown classes as a class and computes the F1 score.}
    \begin{tabular}{l|c|cccc|c}
    \toprule
     Models     & \multicolumn{1}{c|}{known classes} &  \multicolumn{4}{c}{unknown classes} \\
          \cline{3-7}
          & F1-micro &     horse (validation)   &   ship    & truck &  mean   & \multicolumn{1}{c}{one unknown}  \\
          \midrule
    OpenMax &   \textbf{0.695} &    -      &    -   &   -   & - & 0.269  \\
    SoftMax &   \textbf{0.692} &  -    &   -    &    - &  -   & 0.270   \\
    DN,Isometric,D=100 & 0.367$\pm$0.041 &	0.210$\pm$0.079	& 0.168$\pm$0.027 &	\textbf{0.441}$\pm$0.073		&		0.273$\pm$0.027 &	0.444$\pm$0.034 \\
    DN,Share,D=10 & \textbf{0.571}$\pm$0.028	& 0.226$\pm$0.006 &	0.176$\pm$0.105 &	0.095$\pm$0.014		&		0.166$\pm$0.037 &	0.377$\pm$0.019 \\
    DN,Share,D=50 & 0.298$\pm$0.047	& \textbf{0.378}$\pm$0.033 &	\textbf{0.457}$\pm$0.003 &	0.123$\pm$0.003		&		0.319$\pm$0.011 &	0.435$\pm$0.010 \\
    DN,ShareDiag,D=100 & 0.187$\pm$0.029 &	0.253$\pm$0.022	& 0.308$\pm$0.017 &	0.359$\pm$0.024		&		0.307$\pm$0.011 &	\textbf{0.469}$\pm$0.008 \\
    DN,ShareDiag,D=50 & 0.133$\pm$0.054	 & 0.357$\pm$0.022 &	0.401$\pm$0.008 &	0.249$\pm$0.094		&		\textbf{0.335}$\pm$0.029 &	0.432$\pm$0.016 \\
    \bottomrule
    \end{tabular}%
  \label{tab:cifar10}%
\end{table*}%

\section{Experiments}

\subsection{Datasets}
We test the proposed Distribution Networks on two image datasets (MNIST \cite{lecun1998gradient} and CIFAR10 \cite{Krizhevsky09}).
\begin{itemize}
\item MNIST\footnote{http://yann.lecun.com/exdb/mnist/}. We randomly split the provided training set that contains 60000 images into our training set (50000 images) and validation set (10000 images), the provided test set is used for testing, the known classes are 0-6, unknown classes are 7-9.
\item CIFAR10\footnote{https://www.cs.toronto.edu/~kriz/cifar.html}. We randomly split the provided training set that contains 50000 images into our training set (40000 images) and validation set (10000 images), the provided test set (10000 images) is used for testing, the known classes consist of \textit{plane, car, bird, cat, deer, dog and frog}, and the unknown classes consist of \textit{horse, ship and truck}.
%\item Ohsumed corpus\footnote{http://disi.unitn.it/moschitti/corpora.htm}. Each document in Ohsumed dataset has one or more associated categories from 23 disease types. As we focus on single-label classification, the documents belonging to multiple categories are excluded \cite{yao2017incorporating}. We also eliminated classes that have less than 250 samples and Class C23 (i.e., Pathological Conditions, Signs and Symptoms) that is a general disease (different classification criterion with others). This preprocessing resulted in 16 classes with 15389 documents, we randomly split the documents into training (9848), validation (2463) and test (3078) set. We set 13 known classes (\textit{C1, C4, C5, C6, C8, C10, C11, C12, C13, C14, C15, C16, C17}) and 3 unknown classes (\textit{C18, C20, C21}).
\end{itemize}

We also tested other different known-unknown class splits with 3 unknown classes and one of the 3 unknown classes is used for validation. Their results did not show much differences.

\subsection{Settings}
In our experiments, we assume each class follows a Gaussian distribution in the latent space. We have explored 3 types of settings for Gaussian distributions of different classes according to the covariance matrix $\Sigma$.
\begin{itemize}
\item Shared Isometric setting. This assumes that all the classes share a diagonal covariance matrix with equal diagonal elements for their Gaussian distributions. That is, there is only one parameter $\sigma$ related to the covariance matrix  with form $\Sigma=diag(\sigma,\cdots,\sigma)$ for all the distributions.
\item Isometric setting. This setting assumes that each class has an independent diagonal matrix with equal diagonal elements. The isometric setting has a parameter $\sigma_k$ to describe the variance of each class $k$, thus there are $l$ parameters for $l$ classes in the training set.
\item  Shared Diagonal setting. This setting assumes that all the classes share a diagonal covariance matrix for their Gaussian distributions where the diagonal elements do not need to be euqal. Thus the share diagonal settings will have $D$ parameters corresponding to $D$ features.
\end{itemize}

We have also explored the diagonal setting that assumes an independent diagonal covariance matrix for each class, but did not achieve as good results as the other settings in our experiments. This setting has more parameters than the above settings, and should be unnecessarily complex for our experiments. In the following experiments, we explored different latent space dimensions $D\in\{10,50,100\}$ for MNIST dataset and CIFAR10 dataset. Since we assume the latent features are independent, the posterior distribution of each feature can be estimated by Eq. \ref{eq:posterior} and \ref{eq:MAPestimation} independently, in this situation, the dimension $D=1$ for each feature distribution.

As for the mapping network, we use a number of bottom feature layers of random initialized VGG models \cite{simonyan2014very} with batch normalization \cite{ioffe2015batch} for OSL on MNIST dataset and CIFAR10 dataset. The VGG network structure is appended with a fully-connected layer with a $D$-dimensional output. The number of VGG layers is adjusted according to the size of input images to ensure the output size. All of our models were trained via SGD with Adam \cite{kingma2014adam} with learning rate 0.001. The training epoch was set 50. Model selection $\lambda_i$ were selected so that sum of weight of known classes and that of unknown classes are equal, e.g., $\lambda_i=1/3$ for the 3 unknown classes and  $\lambda_i=1/7$ for the 7 known classes.
% We made our source code publicly available at \url{https://github.com/mocherson/Distribution-network}

\subsection{Performances Evaluation}
With the models trained on the training set, we select a model based on the performances on the validation set. We test the selected model on a held-out test set. Since the test process on open set is order-sensitive in our setting, we run the test process 10 times for each model and average the performances. In this setting, the training set discards the unknown classes and only contains the images in the known classes, and the validation set considers all the known classes and one unknown class and sets aside the other unknown classes. The classification performance of known classes is evaluated by the F1 scores. To evaluate the classification performance of an unknown class $k$, we find the created novel class to which most of samples from class $k$ are classified and consider it as the class matched with class $k$, and compute its F1 score.

We compare our model with the OpenMax method \cite{bendale2016towards} as well as the SoftMax method on open set setting with a threshold. We apply the same network structures with output vector size equal to the number of known classes. for OpenMax and SoftMax methods, and report the highest F1 scores by tunning the thresholds. Due to the relative low dimension of MNIST dataset, we also evaluated some other open set recognition algorithms on MNIST dataset (in Table \ref{tab:mnist}). The F1 scores of different models on MNIST and CIFAR10 are shown in Table \ref{tab:mnist} and \ref{tab:cifar10}, respectively.

From Table \ref{tab:mnist} for MNIST dataset, as for known classes recognition in the open set environment, settings ``DN, Share, D=50'', ``DN, Share, D=100'' and ``DN, ShareDiag, D=10'' can outperform OpenMax and other baselines significantly, most of the DN settings can achieve comparable results to baselines. As for the unknown classes, if regarding all unknown classes as one, most of the DN settings can achieve F1 scores over 0.9 which is much better than OpenMax and SoftMax; if considering unknown classes separately, class 7, 8 and 9 can achieve F1 score 0.768, 0.691 and 0.803 respectively, and the mean F1 score can achieve 0.707. This results are very promising given that no samples of these classes are used for training. For CIFAR10, in Table \ref{tab:cifar10}, though DN cannot achieve a better performance than OpenMax and SoftMax for known class recognition, it can recognize unknown classes much better than OpenMax and SoftMax.
%For Ohsumed, in Table \ref{tab:ohsumed}, most of the DN settings can outperform OpenMax and SoftMax for both known class recognition and unknown class discovery. Through the above analysis on the three datasets consisting of both images and documents, we can see that Distribution Networks are able to model novel classes of OSL while maintaining the classification performance of known classes.

\begin{figure}[t]
  \centering
  \subfloat[DN on MNIST]{\includegraphics[width=.5\columnwidth,page=1]{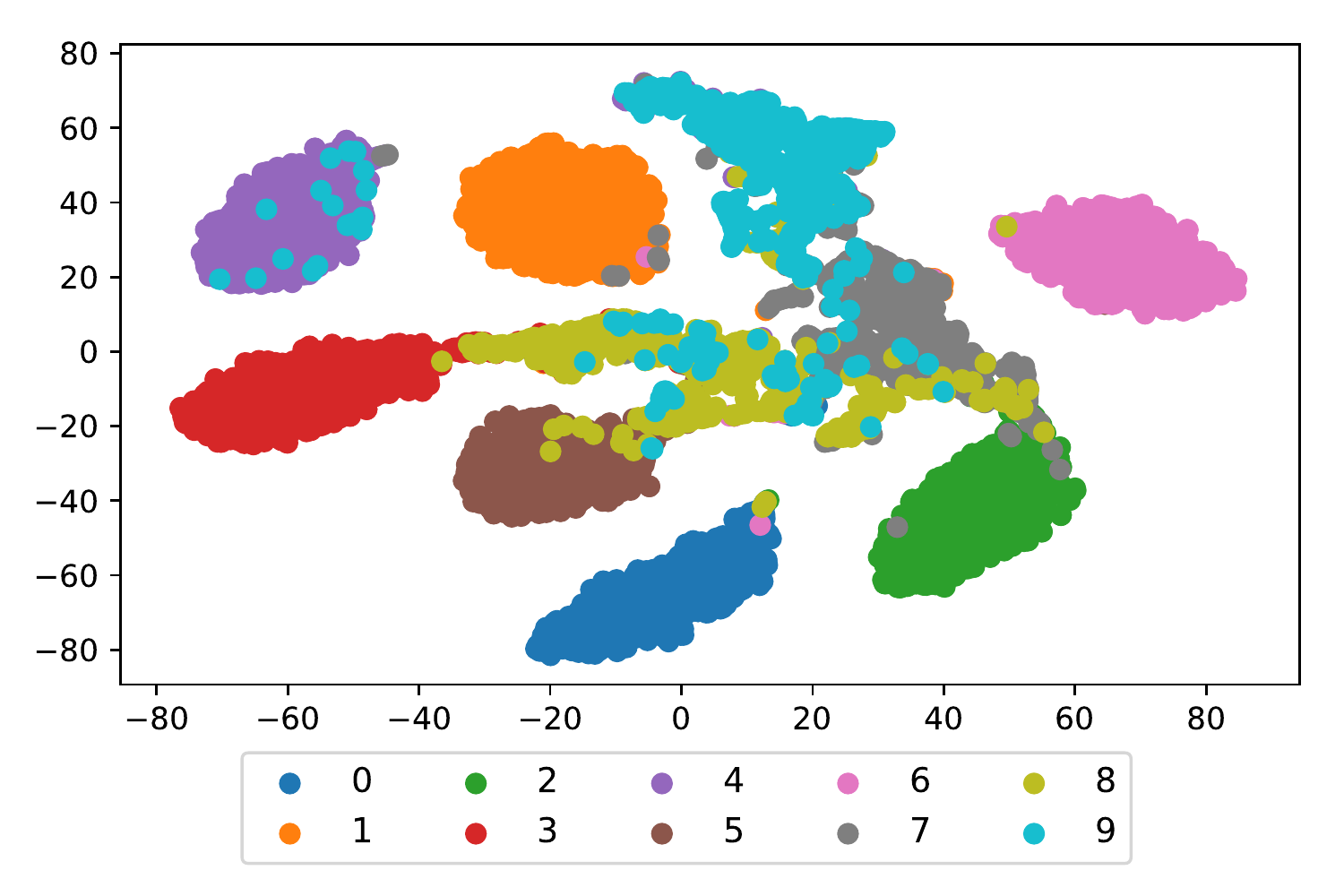}}
  \subfloat[DN on CIFAR10]{\includegraphics[width=.5\columnwidth,page=1]{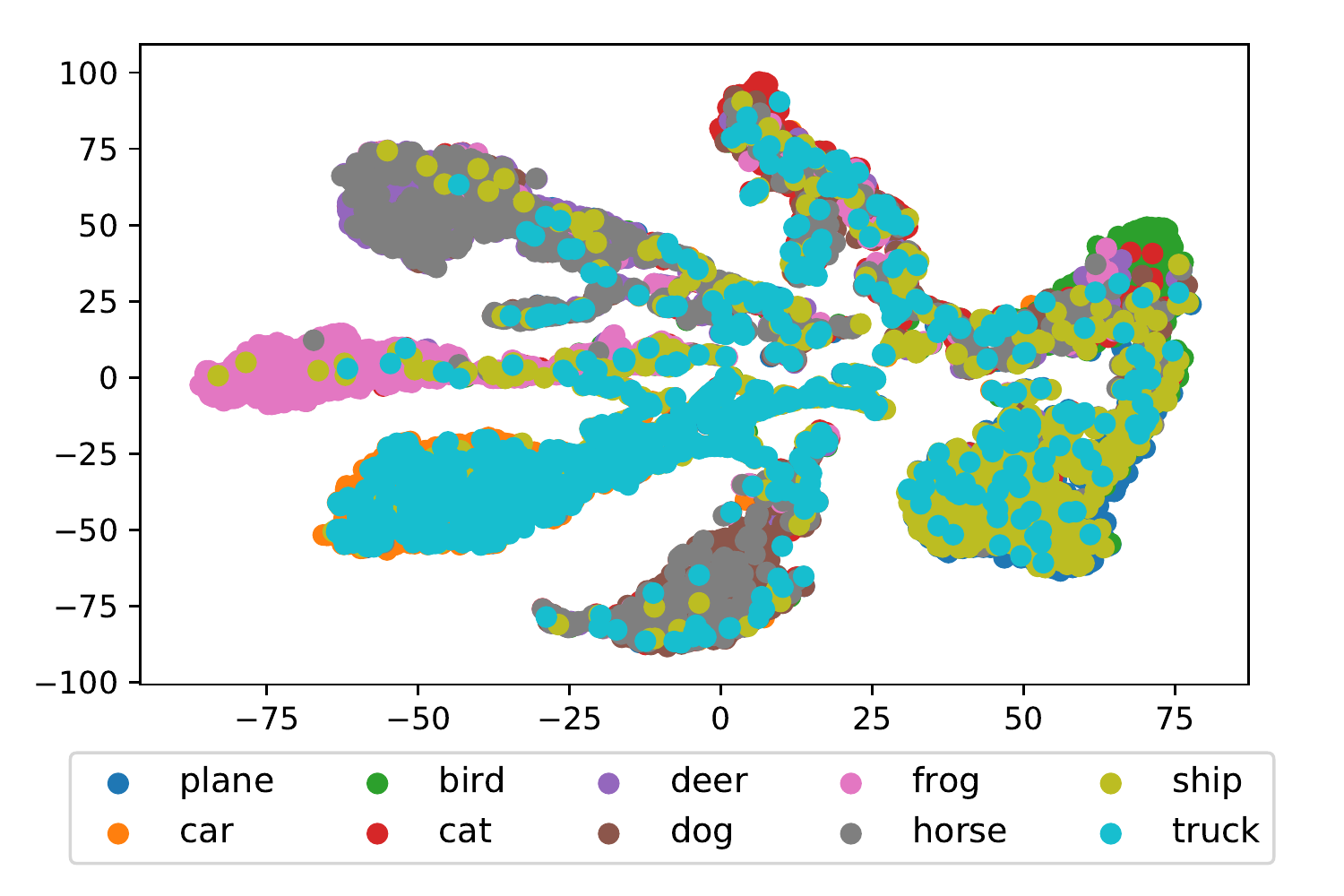}}

  \subfloat[AV on MNIST]{\includegraphics[width=.5\columnwidth,page=1]{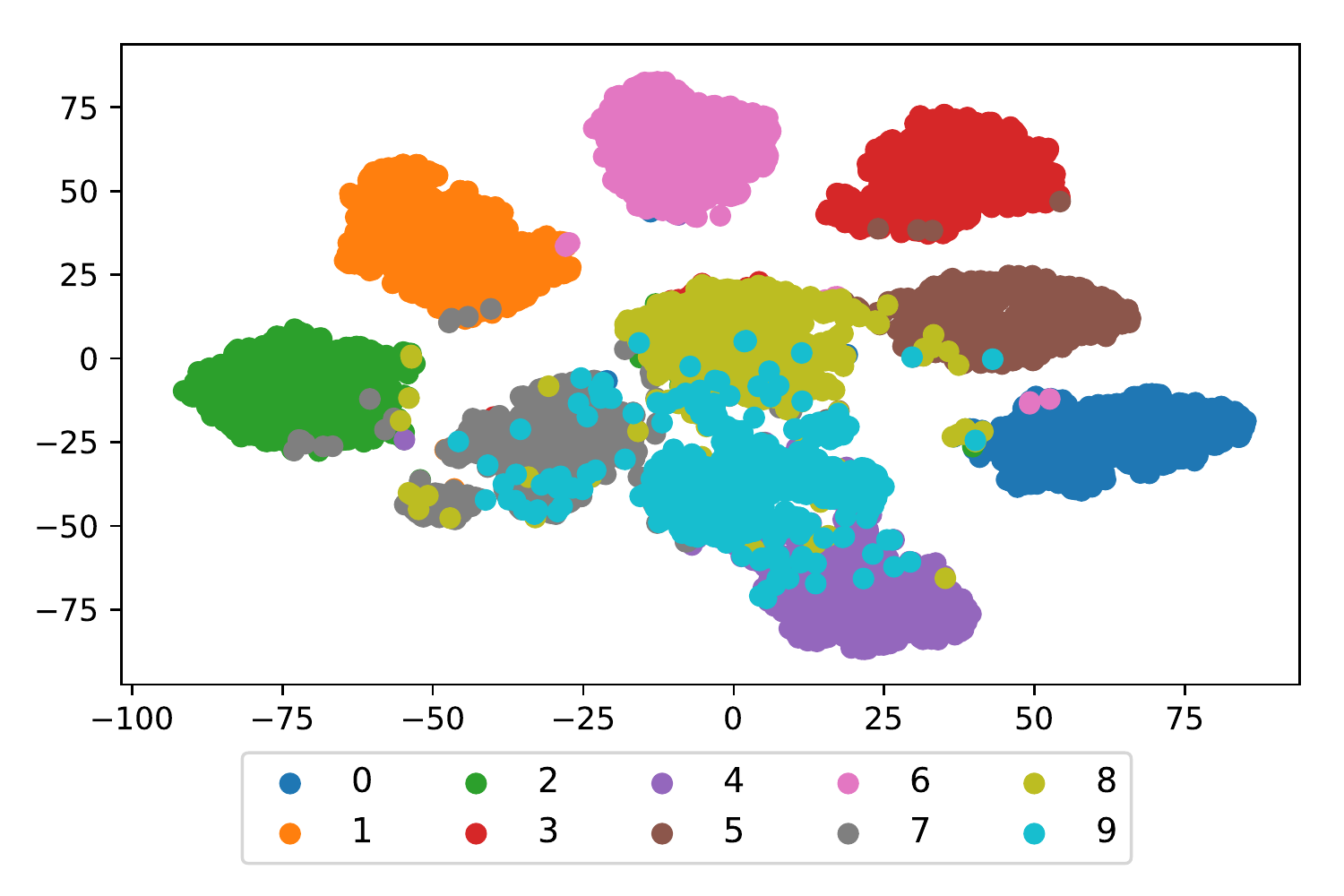}}
  \subfloat[AV on CIFAR10]{\includegraphics[width=.5\columnwidth,page=1]{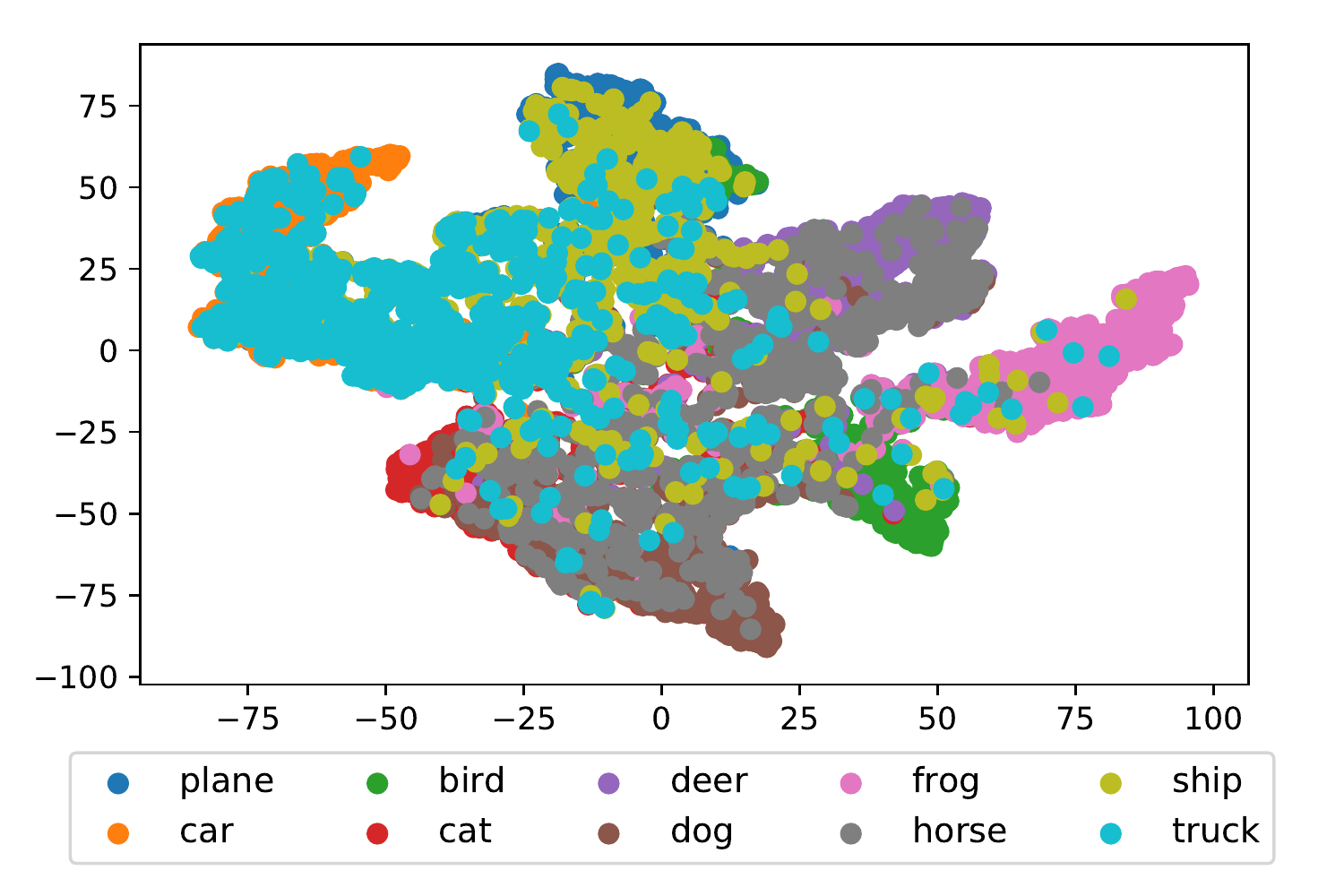}}
  \caption{t-SNE visualization of the latent embeddings on test set with DN and activation vectors (AV) of a general deep network with the same settings. The unknown classes are $\{7,8,9\}$ for MNIST, and $\{horse, ship, truck\}$ for CIFAR10. The DN settings are ``DN, Share, D=50'' for MNIST and ``DN, Share, D=10'' for CIFAR10.}
  \label{fig:cluster}
\end{figure}

To better understand distribution networks, in Figure \ref{fig:cluster}, we show the t-SNE visualization \cite{maaten2008visualizing} of the latent embeddings of test set learned by DN, as well as the activation vectors (AV) \cite{bendale2016towards} learned by a general deep network with the same settings. The unknown classes are $\{7,8,9\}$ for MNIST dataset, $\{horse, ship, truck\}$ for CIFAR10 dataset. We observe that the samples in MNIST can be tightly clustered to the corresponding classes by both DN and AV, even for unknown classes. For CIFAR10 dataset, we can see that DN can cluster the test data much better than AV, the unknown classes can be well discriminated between each other. There are some known classes are indeed visually similar with the unknown classes, so that DN can not totally discriminate them, e.g., \textit{horse vs. deer, truck vs. car}.

\begin{figure}[t]
  \centering
  \subfloat[Known classes]{\includegraphics[width=.5\columnwidth,page=1]{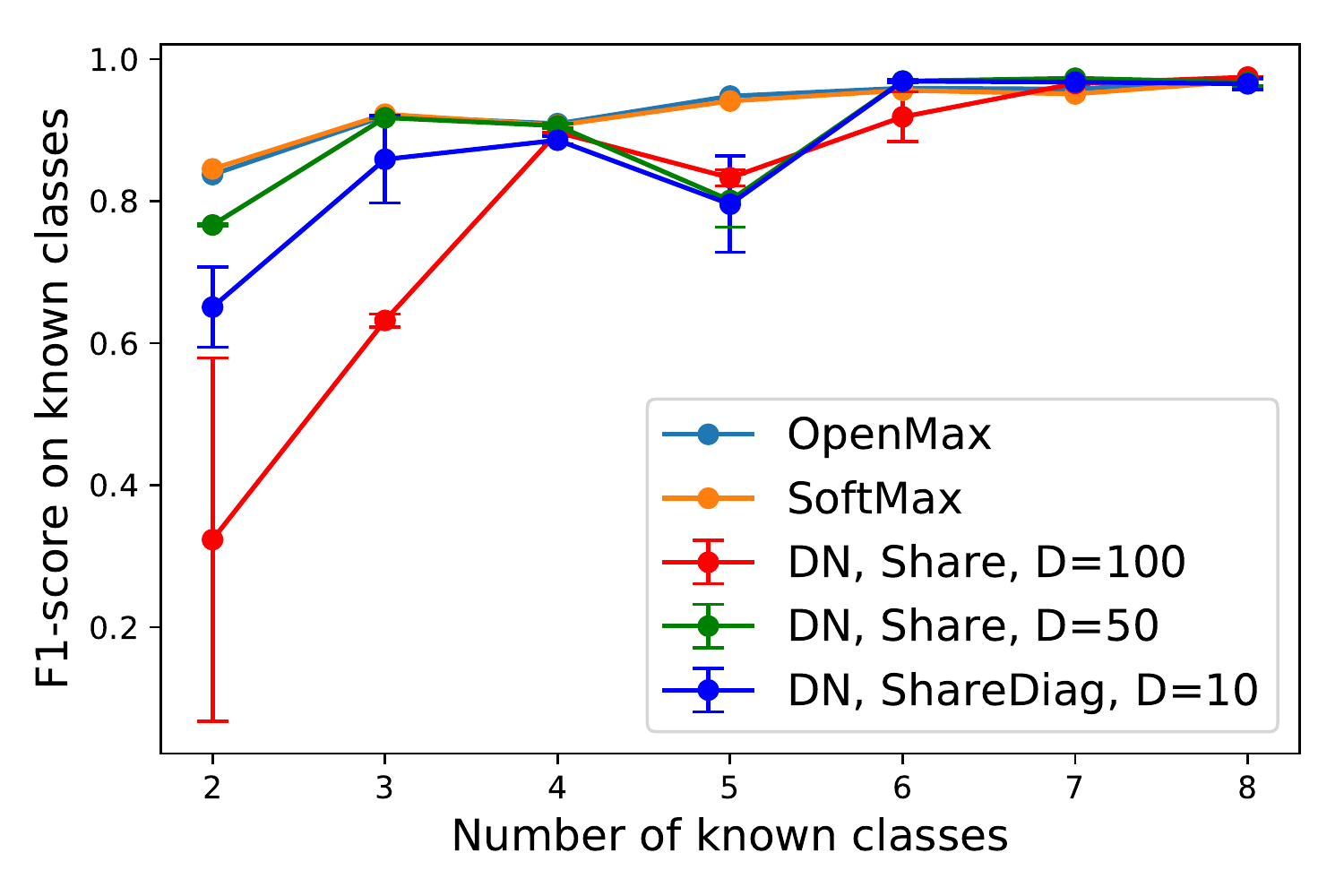}
  \label{fig:opennessknown}}
  \subfloat[Unknown classes]{\includegraphics[width=.5\columnwidth,page=1]{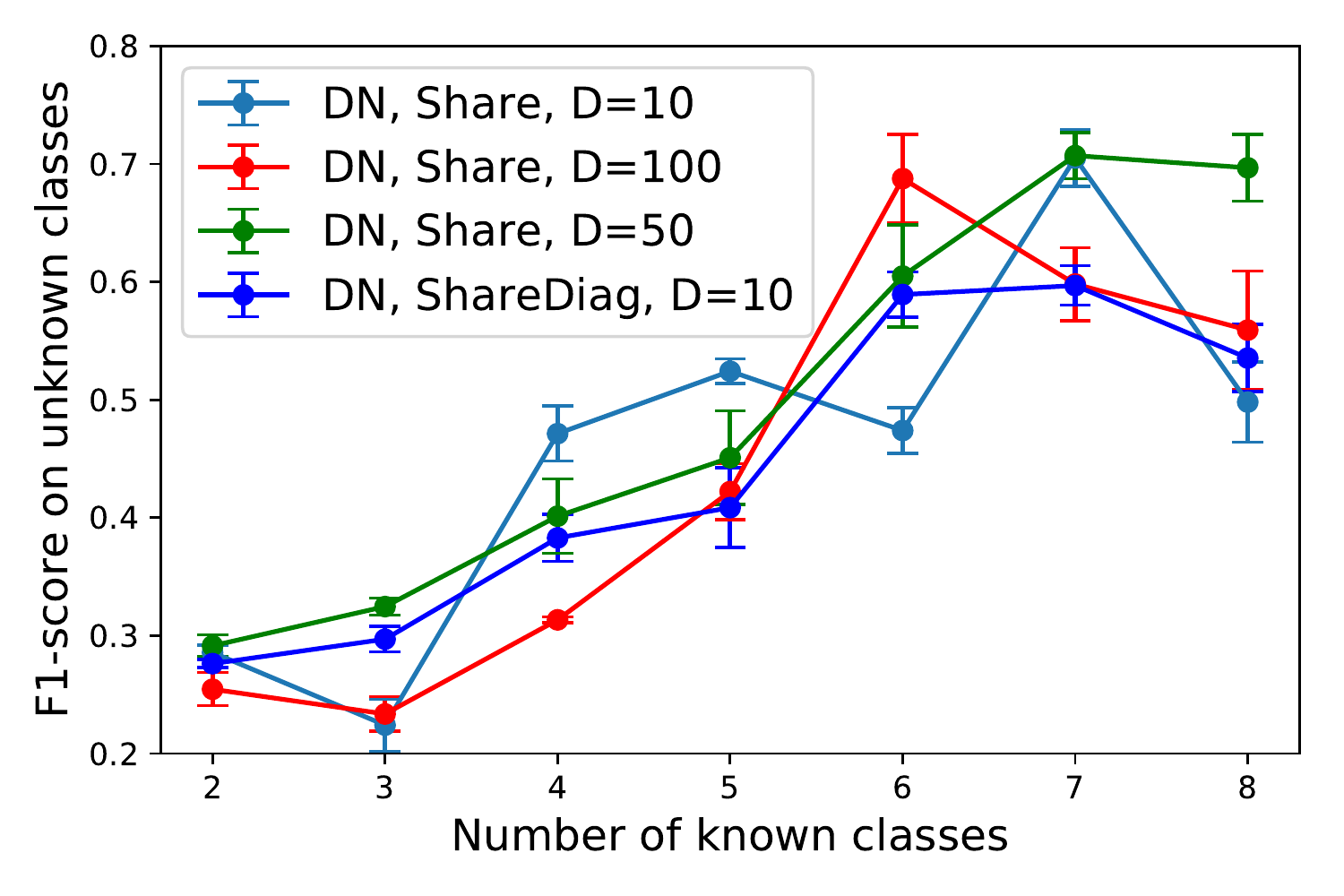}
  \label{fig:opennessunknown}}
  \caption{The influence of openness on the classification performance on MNIST dataset.}
  \label{fig:openness}
\end{figure}

\textbf{Openness analysis.}
Intuitively, making more known classes available in the training process should improve the classification performance. We will analyse the influence of openness on classification performance in this paragraph. For a open set problem, Scheirer et al. \cite{scheirer2013toward,scheirer2014probability} defined the degree of openness as $openness=1-\sqrt{2\times \mathcal{C}_T/(\mathcal{C}_R+\mathcal{C}_E)}$, where $\mathcal{C}_T$, $\mathcal{C}_R$, and $\mathcal{C}_E$ are the number of training classes, the number of classes to be identify, and the number of testing classes, respectively. In our settings, we want to identify all the classes in the test class, thus $\mathcal{C}_R=\mathcal{C}_E=10$ for MNIST dataset and CIFAR10 dataset, openness is only related to the number of training classes. We change the number of training classes to train the distribution network, and test its performance on the test set to see the openness influence. We plot the recognition performance of known classes and unknown classes varied with the number of known classes on MNIST dataset in Figure \ref{fig:openness}. In Figure \ref{fig:opennessknown}, we can see that, with the increasing number of known classes, the recognition performance of known classes increases roughly, and that OpenMax and SoftMax perform better than DN when the number of known classes is small. The recognition performance of unknown classes also increases roughly as the number of known classes increases as seen in Figure \ref{fig:opennessunknown}. The results are consistent with the intuition that training with more known classes will improve the classification performance for both known classes and unknown classes.

\section{Conclusion}
We proposed Distribution Networks for open set learning tasks based on the idea that different classes by the same classification criterion can be mapped to a latent space where the classes follow different distributions from the same distribution family. Thus through training a distribution network to map known classes to known distributions, we also mapped the unknown classes to distributions with known types, enabling the detection and modeling of the unknown classes. Our experiments on image datasets MNIST and CIFAR10 validated our expectation that the unknown classes could be discovered and modeled during testing even there were no training samples for these classes. Additionally, we raised the problem of modeling novel classes for open set learning, modeling novel classes for the subsequent classification is of great significance for us to discover new patterns and new knowledge, we tackled the problem with Distribution Networks. Further, we recommended using the incremental testing settings for open set learning. The limitation of our model is that the classes were assumed to be from the same distribution family through a certain mapping from original feature space to a lower latent dimensional feature space. The model may fail if the assumption does not hold. In addition, the model cannot be extended to multi-label cases, because a distribution network is a one-to-one mapping, if an image contains multiple objects that are very different, classification of the image would fail. Augmenting the model to fit multi-label open set classification problem is a part of our future work.

{\small
\bibliographystyle{ieee}
\bibliography{openset}

\begin{thebibliography}{10}\itemsep=-1pt

\bibitem{akova2010machine}
F.~Akova, M.~Dundar, V.~J. Davisson, E.~D. Hirleman, A.~K. Bhunia, J.~P.
  Robinson, and B.~Rajwa.
\newblock A machine-learning approach to detecting unknown bacterial serovars.
\newblock {\em Statistical Analysis and Data Mining: The ASA Data Science
  Journal}, 3(5):289--301, 2010.

\bibitem{al2016recurring}
T.~Al-Khateeb, M.~M. Masud, K.~M. Al-Naami, S.~E. Seker, A.~M. Mustafa,
  L.~Khan, Z.~Trabelsi, C.~Aggarwal, and J.~Han.
\newblock Recurring and novel class detection using class-based ensemble for
  evolving data stream.
\newblock {\em IEEE Transactions on Knowledge and Data Engineering},
  28(10):2752--2764, 2016.

\bibitem{bendale2015towards}
A.~Bendale and T.~Boult.
\newblock Towards open world recognition.
\newblock {\em 2015 Ieee Conference on Computer Vision and Pattern Recognition
  (Cvpr)}, pages 1893--1902, 2015.

\bibitem{bendale2016towards}
A.~Bendale and T.~E. Boult.
\newblock Towards open set deep networks.
\newblock {\em 2016 Ieee Conference on Computer Vision and Pattern Recognition
  (Cvpr)}, pages 1563--1572, 2016.

\bibitem{cardoso2015bounded}
D.~O. Cardoso, F.~Fran{\c{c}}a, and J.~Gama.
\newblock A bounded neural network for open set recognition.
\newblock In {\em Neural Networks (IJCNN), 2015 International Joint Conference
  on}, pages 1--7. IEEE, 2015.

\bibitem{da2014learning}
Q.~Da, Y.~Yu, and Z.-H. Zhou.
\newblock Learning with augmented class by exploiting unlabeled data.
\newblock In {\em Twenty-Eighth AAAI Conference on Artificial Intelligence},
  2014.

\bibitem{dietterich2017steps}
T.~G. Dietterich.
\newblock Steps toward robust artificial intelligence.
\newblock {\em AI Magazine}, 38(3):3--24, 2017.

\bibitem{dundar2012bayesian}
M.~Dundar, F.~Akova, Y.~Qi, and B.~Rajwa.
\newblock Bayesian nonexhaustive learning for online discovery and modeling of
  emerging classes.
\newblock In {\em Proceedings of the 29th International Coference on
  International Conference on Machine Learning}, pages 99--106. Omnipress,
  2012.

\bibitem{Dundar2009Learning}
M.~M. Dundar, E.~D. Hirleman, A.~K. Bhunia, J.~P. Robinson, and B.~Rajwa.
\newblock Learning with a non-exhaustive training dataset: A case study:
  Detection of bacteria cultures using optical-scattering technology.
\newblock In {\em Proceedings of the 15th ACM SIGKDD International Conference
  on Knowledge Discovery and Data Mining}, KDD '09, pages 279--288, New York,
  NY, USA, 2009. ACM.

\bibitem{Ge2017GenerativeOF}
Z.~Ge, S.~Demyanov, and R.~Garnavi.
\newblock Generative openmax for multi-class open set classification.
\newblock {\em CoRR}, abs/1707.07418, 2017.

\bibitem{gepperth2016bio}
A.~Gepperth and C.~Karaoguz.
\newblock A bio-inspired incremental learning architecture for applied
  perceptual problems.
\newblock {\em Cognitive Computation}, 8(5):924--934, 2016.

\bibitem{gunther2017toward}
M.~G{\"u}nther, S.~Cruz, E.~M. Rudd, and T.~E. Boult.
\newblock Toward open-set face recognition.
\newblock In {\em Conference on Computer Vision and Pattern Recognition (CVPR)
  Workshops. IEEE}, 2017.

\bibitem{haque2016sand}
A.~Haque, L.~Khan, and M.~Baron.
\newblock Sand: Semi-supervised adaptive novel class detection and
  classification over data stream.
\newblock In {\em AAAI}, pages 1652--1658, 2016.

\bibitem{hassen2018learning}
M.~Hassen and P.~K. Chan.
\newblock Learning a neural-network-based representation for open set
  recognition.
\newblock {\em arXiv preprint arXiv:1802.04365}, 2018.

\bibitem{ioffe2015batch}
S.~Ioffe and C.~Szegedy.
\newblock Batch normalization: accelerating deep network training by reducing
  internal covariate shift.
\newblock In {\em Proceedings of the 32nd International Conference on
  International Conference on Machine Learning-Volume 37}, pages 448--456.
  JMLR. org, 2015.

\bibitem{jain2014multi}
L.~P. Jain, W.~J. Scheirer, and T.~E. Boult.
\newblock Multi-class open set recognition using probability of inclusion.
\newblock {\em Computer Vision - Eccv 2014, Pt Iii}, 8691:393--409, 2014.

\bibitem{kemker2018fearnet}
R.~Kemker and C.~Kanan.
\newblock Fearnet: Brain-inspired model for incremental learning.
\newblock In {\em International Conference on Learning Representations}, 2018.

\bibitem{kingma2014adam}
D.~P. Kingma and J.~Ba.
\newblock Adam: A method for stochastic optimization.
\newblock In {\em International Conference on Learning Representations}, 2015.

\bibitem{koch2015siamese}
G.~Koch, R.~Zemel, and R.~Salakhutdinov.
\newblock Siamese neural networks for one-shot image recognition.
\newblock In {\em ICML Deep Learning Workshop}, volume~2, 2015.

\bibitem{kodirov2017semantic}
E.~Kodirov, T.~Xiang, and S.~G. Gong.
\newblock Semantic autoencoder for zero-shot learning.
\newblock {\em 30th Ieee Conference on Computer Vision and Pattern Recognition
  (Cvpr 2017)}, pages 4447--4456, 2017.

\bibitem{Krizhevsky09}
A.~Krizhevsky and G.~Hinton.
\newblock Learning multiple layers of features from tiny images.
\newblock {\em Master's thesis, Department of Computer Science, University of
  Toronto}, 2009.

\bibitem{lecun1998gradient}
Y.~LeCun, L.~Bottou, Y.~Bengio, and P.~Haffner.
\newblock Gradient-based learning applied to document recognition.
\newblock {\em Proceedings of the IEEE}, 86(11):2278--2324, 1998.

\bibitem{maaten2008visualizing}
L.~v.~d. Maaten and G.~Hinton.
\newblock Visualizing data using t-sne.
\newblock {\em Journal of machine learning research}, 9(Nov):2579--2605, 2008.

\bibitem{Masud2011Classification}
M.~M. Masud, J.~Gao, L.~Khan, J.~W. Han, and B.~Thuraisingham.
\newblock Classification and novel class detection in concept-drifting data
  streams under time constraints.
\newblock {\em Ieee Transactions on Knowledge and Data Engineering},
  23(6):859--874, 2011.

\bibitem{mu2017streaming}
X.~Mu, F.~Zhu, J.~Du, E.-P. Lim, and Z.-H. Zhou.
\newblock Streaming classification with emerging new class by class matrix
  sketching.
\newblock In {\em AAAI}, pages 2373--2379, 2017.

\bibitem{murphy2012machine}
K.~P. Murphy.
\newblock {\em Machine Learning: A Probabilistic Perspective}.
\newblock MIT Press, 2012.

\bibitem{neal2018open}
L.~Neal, M.~Olson, X.~Fern, W.-K. Wong, and F.~Li.
\newblock Open set learning with counterfactual images.
\newblock In {\em Proceedings of the European Conference on Computer Vision
  (ECCV)}, pages 613--628, 2018.

\bibitem{palatucci2009zero}
M.~Palatucci, D.~Pomerleau, G.~E. Hinton, and T.~M. Mitchell.
\newblock Zero-shot learning with semantic output codes.
\newblock In {\em Advances in neural information processing systems}, pages
  1410--1418, 2009.

\bibitem{rebuffi2017icarl}
S.-A. Rebuffi, A.~Kolesnikov, G.~Sperl, and C.~H. Lampert.
\newblock icarl: Incremental classifier and representation learning.
\newblock In {\em Proc. CVPR}, 2017.

\bibitem{scheirer2014probability}
W.~J. Scheirer, L.~P. Jain, and T.~E. Boult.
\newblock Probability models for open set recognition.
\newblock {\em IEEE transactions on pattern analysis and machine intelligence},
  36(11):2317--2324, 2014.

\bibitem{scheirer2013toward}
W.~J. Scheirer, A.~D. Rocha, A.~Sapkota, and T.~E. Boult.
\newblock Toward open set recognition.
\newblock {\em Ieee Transactions on Pattern Analysis and Machine Intelligence},
  35(7):1757--1772, 2013.

\bibitem{simonyan2014very}
K.~Simonyan and A.~Zisserman.
\newblock Very deep convolutional networks for large-scale image recognition.
\newblock In {\em International Conference on Learning Representations}, 2015.

\bibitem{snell2017prototypical}
J.~Snell, K.~Swersky, and R.~Zemel.
\newblock Prototypical networks for few-shot learning.
\newblock In {\em Advances in Neural Information Processing Systems}, pages
  4080--4090, 2017.

\bibitem{socher2013zero}
R.~Socher, M.~Ganjoo, C.~D. Manning, and A.~Ng.
\newblock Zero-shot learning through cross-modal transfer.
\newblock In {\em Advances in neural information processing systems}, pages
  935--943, 2013.

\bibitem{sutskever2013importance}
I.~Sutskever, J.~Martens, G.~Dahl, and G.~Hinton.
\newblock On the importance of initialization and momentum in deep learning.
\newblock In {\em International conference on machine learning}, pages
  1139--1147, 2013.

\bibitem{vinyals2016matching}
O.~Vinyals, C.~Blundell, T.~Lillicrap, and D.~Wierstra.
\newblock Matching networks for one shot learning.
\newblock In {\em Advances in Neural Information Processing Systems}, pages
  3630--3638, 2016.

\bibitem{zhang2017learning}
L.~Zhang, T.~Xiang, and S.~G. Gong.
\newblock Learning a deep embedding model for zero-shot learning.
\newblock {\em 30th Ieee Conference on Computer Vision and Pattern Recognition
  (Cvpr 2017)}, pages 3010--3019, 2017.

\end{thebibliography}
}

\end{document}